\documentclass[10pt, a4paper]{article}

\usepackage[final]{lrec-coling2024} 

\usepackage{color}
\usepackage{xcolor}
\usepackage{tikz}
\usepackage{hyperref}

\newcommand{\rev}[1]{#1}
\newcommand{\grayc}[1]{{\color{gray} #1}}
\newcommand{\newrev}[1]{{#1}}
\newcommand{\newnewrev}[1]{{\color{black} #1}}
\usepackage{stmaryrd}
\newcommand{\improve}[1]{\color{blue!20!black!30!green}($\downarrow${#1})}
\usetikzlibrary{arrows}
\usepackage{colonequals}
\usepackage{booktabs}
\usepackage{subcaption}
\usepackage{multirow}
\usepackage{amsmath}
\usepackage{placeins}
\usepackage{amssymb}
\usepackage[normalem]{ulem}

\usepackage{physics}

\usepackage{color}
\usepackage{colortbl}
\usepackage{tcolorbox}

\usepackage{amsmath,amsfonts,bm}

\def\eqref#1{equation~\ref{#1}}

\def\1{\bm{1}}
\newcommand{\train}{\mathcal{D}}
\newcommand{\valid}{\mathcal{D_{\mathrm{valid}}}}
\newcommand{\test}{\mathcal{D_{\mathrm{test}}}}

\def\rmA{{\mathbf{A}}}

\def\rmK{{\mathbf{K}}}

\def\rmM{{\mathbf{M}}}

\def\rmQ{{\mathbf{Q}}}

\def\rmV{{\mathbf{V}}}

\def\vtheta{{\bm{\theta}}}

\DeclareMathAlphabet{\mathsfit}{\encodingdefault}{\sfdefault}{m}{sl}
\SetMathAlphabet{\mathsfit}{bold}{\encodingdefault}{\sfdefault}{bx}{n}

\newcommand{\R}{\mathbb{R}}

\newcommand{\softmax}{\mathrm{softmax}}

\title{Data-Informed Global Sparseness in Attention Mechanisms \\
for Deep Neural Networks}

\name{Ileana Rugina\textsuperscript{*, \rm 1}, Rumen Dangovski\textsuperscript{*, \rm 1}, Li Jing\textsuperscript{ \rm 1}, Preslav Nakov\textsuperscript{\rm 2}, Marin Soljačić\textsuperscript{ \rm 1}} 

\address{{\rm (*)} equal contribution \\
         \textsuperscript{\rm 1} Massachusetts Institute of Technology \\
         \textsuperscript{\rm 2} Mohamed bin Zayed University of Artifcial Intelligence, UAE\\
         \{irugina, rumenrd, ljing, soljacic\}@mit.edu \\
         preslav.nakov@mbzuai.ac.ae
         }

\abstract{
Attention mechanisms play a crucial role in the neural revolution of Natural Language Processing (NLP). With the growth of attention-based models, several pruning techniques have been developed to identify and exploit sparseness, making these models more efficient. Most efforts focus on hard-coding attention patterns or pruning attention weights based on training data. We propose Attention Pruning (AP), a framework that observes attention patterns in a fixed dataset and generates a global sparseness mask. AP saves 90\% of attention computation for language modeling and about 50\% for machine translation and GLUE tasks, maintaining result quality. Our method reveals important distinctions between self- and cross-attention patterns, guiding future NLP research. \newnewrev{Our framework can reduce both latency and memory requirements for any attention-based model, aiding in the development of improved models for existing or new NLP applications. We have demonstrated this with encoder and autoregressive transformer models using Triton GPU kernels and make our code publicly available at  \href{https://github.com/irugina/AP}{https://github.com/irugina/AP}}
 \\ \newline  }

\begin{document}

\maketitleabstract

\section{Introduction}

Given enough computational power, the scalability of the attention mechanism~\citep{bahdanau2015,hermann2015,transformer} will allow for building ever larger Natural Language Processing (NLP) models with billions of parameters \citep{shoeybi2019megatronlm, Radford2019LanguageMA, raffel2019exploring, brown2020language,chowdhery2022palm,smith2022using,fedus2022switch,du2022glam,anil2023palm}.

While impressive, these advances also pose a responsibility to the Natural Language Processing (NLP) community to interpret the behavior of the hundreds of attention heads in a single model, and potentially to reduce the number of computations. Responding to this challenge, previous work has taken pioneering steps to discover and to explain the sparseness in the attention patters~\cite{vig2019,Clark2019WhatDB,kovaleva-etal-2019-revealing,yeh2023attentionviz,ruscio2023attention,kobayashi2023feed,biderman2023pythia,zhang2023better,li2023does}. Here, we argue that as the number of heads grows in the range of thousands, automatic measures would be needed to discover and to impose sparseness to such models.

In this paper we introduce a simple task-agnostic data-informed pruning method for attention mechanisms: \emph{Attention Pruning}. We train Transformer-based models and we analyze the \emph{global} observed attention patterns, averaged over all input sequences in the train set, in order to identify and to remove weak connections between input tokens.

\begin{figure}[t]
    \centering    \includegraphics[width=\linewidth]{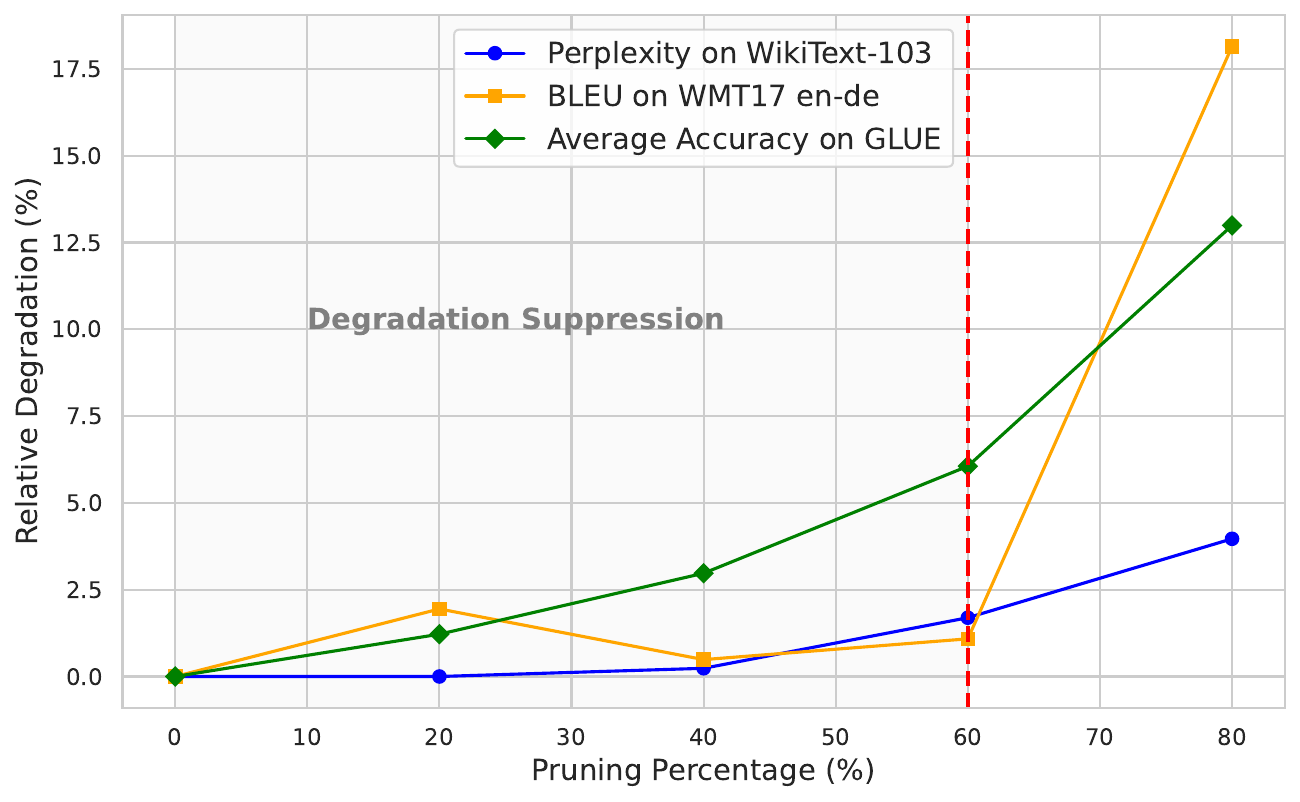}
    \caption{\rev{Attention pruning maintains performance and reduced attention computations. Pruning can often enable more efficient and interpretable models with only a modest decrease in performance. }}
    \label{fig:visualization_ap}
\end{figure}

Following \citet{lottery}, we then retrain these models, enforcing sparseness through masking, and we demonstrate that attention mechanisms incorporate extraneous connections between the input tokens: we obtain comparable 
performance while using sparse attention patterns for NLP tasks such as language and sequence-to-sequence (seq2seq) modelling, as well as 
\rev{prediction on GLUE tasks. Figure~\ref{fig:visualization_ap} summarizes the impact of using our pruning method on standard NLP tasks.}

These global sparseness patterns could help improve both interpretability and inference-time computational efficiency for widely-used attention models. Our contributions are as follows:

\begin{itemize}
\item We present a novel pruning method for attention patterns, focusing on the model's computational graph, not attention weights, and showcase both theoretical and empirical gains.
\item Our data-informed, global method prunes attention while retaining information and closely preserving original results.
\item As an application-agnostic method, we investigate pruning impacts on language and seq2seq modeling, including GLUE task predictions.
\item In seq2seq experiments, we examine attention pruning in encoder self-attention, decoder self-attention, and encoder--decoder attention, highlighting key differences.
\end{itemize}

\section{Related Work}
\label{sec:related}

There are several fruitful directions for research focused on improving the computational efficiency and the interpretability of the attention mechanism. Sparseness plays a central role in all of them, as simple attention mechanisms inherently scale quadratically with sequence length and assign non-zero correlation between any two input tokens.

\newrev{One line of research is that of reducing computational complexity of attention using insights provided by empirical observations of patterns. \citet{Child2019GeneratingLS} introduced two sparse matrix factorizations that reduce the computational complexity from $O(N^2)$ to $O(N\sqrt{N})$. \citet{reformer} created a sparse attention mechanism with $O(N\log N)$ computational complexity, which is achieved by using local sensitivity hashing to cluster tokens that should attend to each other and then only computing attention within tokens from the same chunks. More directly related to our work is that of \citet{longformer} and \citet{NEURIPS2020_c8512d14} who looked directly at sparsifying the attention patterns rather than at the underlying  matrix factorization, and reduced the computational complexity of attention from $O(N^2)$ to $O(N)$ using GPU kernels \citep{gray2017}.} A key difference between these approaches and ours is that we do not impose any a priori restrictions on the type of attention patterns we can generate.

\newrev{Another successful approach has been to adapt low-rank matrix approximation methods to simplify attention computations. \citet{xiong2021nystromformer} adapted Nystr\"{o}m's method~\citep{Baker1979}. 
\citet{wang2020linformer} leveraged the 
Johnson–Lindenstrauss lemma to introduce projections and to improve complexity. In contrast to this line of work our contribution is easier to implement because (\emph{i})~we do not require architectural modifications and we only change a few lines of code in practice, and (\emph{ii})~we do not require optimizing the numerical stability of any mathematical methods and we introduce very few and simple hyper-parameters.}

\citet{martins_adapt_sparse} and \citet{martins_sparse-seq2seq} explored directly incorporating sparseness into Transformer models through the choice of activation functions and introduced the $\alpha$-entmax functions. This encompasses both softmax for $\alpha = 1$ and sparsemax (or projections onto the probability simplex) for $\alpha = 2$. For any $\alpha > 1$ $\alpha$-entmax is sparse. \citet{martins_sparse-seq2seq} provided an efficient implementation and experimented with $\alpha = 1.5$. \rev{We leverage global attention masks, rather than creating a sparse attention pattern for each key-value pair, and we manage both to provide quantifiable speed guarantees and to achieve higher sparseness in practice.}

There has been a lot of research on understanding over-parameterization and on developing methods that make BERT models faster and more lightweight~\citep{Ganesh2020CompressingLT}. Previous work has found that different attention heads encode similar patterns and hence these heads are not always all necessary \citep{NIPS2019_9551, kovaleva-etal-2019-revealing, voita2019}, and thus good performance can be achieved by removing entire attention heads at test time. \citet{Sajjad2020PoorMB} pruned entire Transformer layers at a time and again obtained good performance while removing a large percentage of the model's parameters. Our pruning method takes a more fine-grained approach and prunes individual connections rather than entire heads or layers, and thus it could be used in conjunction with the above-mentioned methods.

\section{Attention Pruning}
\label{sec:ap}

\subsection{Scaled Dot-Product Attention}

Attention mechanisms are used to learn connections (correlations) between two sequences of lengths $N$ and $M$, respectively.

Transformer models use the scaled dot-product attention introduced in~\citep{transformer}, which takes as input three matrices: a matrix $\rmQ \in \mathbb{R}^{M \times d_k} $ composed of query vectors $q \in \mathbb{R}^{d_k}$, a matrix $\rmK \in \mathbb{R}^{N\times d_k}$ composed of key vectors $k \in \mathbb{R}^{d_k}$, and a third matrix $\rmV \in \mathbb{R}^{N \times d_v}$, which groups value vectors $v \in \mathbb{R}^{d_v}$.

The scaled dot-product outputs a transformation of the sequence of length $M$ governed by the values associated with other sequence's tokens, as well as the relative strength of the connection between any two tokens from the two sequences:

\begin{equation*} \label{old:new-attention}
\rmA= \softmax\left( \frac{\rmQ \rmK^\top }{\sqrt{d_k}} \right)\rmV
\end{equation*}

The general sequence-to-sequence Transformer model uses three types of attention layers: two self-attention mechanisms for the encoder and for the decoder, respectively, as well as a third encoder--decoder attention, which we will call \emph{cross-attention}. Additionally, each attention layer is a multi-headed attention mechanisms that concatenates several instances of the dot-product attention described above.

\subsection{Our Pruning Algorithm}

\textbf{Tailor attention to the dataset.}
The attention mechanism \citep{transformer} creates a matrix with $NM$ connections where $N$ is the length of a source sequence, i.e.,~its number of tokens, and $M$ is the length of the target sequence. In practice, in order to compute high-level representations of sources and targets, a portion of these connections is ignored by pre-trained models, which could be observed with visualization tools on pre-trained Transformer-based models \citep{vig2019, Clark2019WhatDB, kovaleva-etal-2019-revealing}. More specifically, there are entries in the attention matrix whose values are close to zero \textit{with high probability}, and thus there is no correlation between the corresponding tokens. We would like to prune such connections, thus reducing the noise coming from such entries.

For each possible tuple $(\mathrm{type}, l, m)$, where $\mathrm{type}$ is either a \emph{self-attention encoder}, a \emph{self-attention decoder} or an \emph{encoder--decoder}, $l$ \rev{$\in \{1,\dots L\}$} is the layer in the Transformer, $m$ \rev{$\in \{1,\dots H\}$} is the index of the head within the Transformer layer, we calculate the corresponding \textit{average attention matrix} $\overline{\rmA}^{(\mathrm{type},l,m)}$ from the corresponding attention matrices $\rmA_i^{(\mathrm{type},l,m)}$ generated by source-target pairs in the training set as follows:
\begin{equation} \label{eq:accum}\overline{\rmA}^{(\mathrm{type},l,m)} = \frac{1}{n} \sum_{i=1}^n  \rmA_i^{(\mathrm{type},l,m)} ,
\end{equation}
where $n$ is the size of the training set. The summation is element-wise.

\rev{We introduce a single additional hyper-parameter $p$: the percentage of entries we want to prune in each multi-head attention mechanism. We would prune an entry in the attention matrix if almost all of the source-target pairs in the training dataset are below a threshold $\tau^{(type, l)}$, given by the $p$ percentile of  $\left[\overline{\rmA}^{(\mathrm{type},l,1)}, \dots, \overline{\rmA}^{(\mathrm{type},l,H)}\right]$. In other words, the global percentage $p$ defines a layer-dependent pruning threshold $\tau^{(type, l)}$, which corresponds to the $p$ percentile of all attention entries within that layer, across all heads. The pruning mask $\rmM_p$, which depends on $p$, is computed as follows:

\begin{equation} \label{eq:mask}
\rmM_p^{(\mathrm{type},l,m)} = (-\infty) \cdot \left\llbracket \overline{\rmA}^{(\mathrm{type},l,m)} < \tau^{(type, l)} \right\rrbracket.\footnote{We implement this by adding a large negative value before applying softmax; this value depends on the codebase.}
\end{equation}
}

We use the Iverson bracket notation, where the comparison operation yields a matrix, whose entries are 1 if the comparison is satisfied, and 0 otherwise (using broadcasting for the scalars). 

The mask modifies the calculation of the attention matrix as follows:
\begin{equation*} 
\label{eq:new-attention}
\widetilde{\rmA}_{i,p}^{(t)} = \softmax\left( \frac{\rmQ_i^{(t)} \rmK_i^{(t)\top} + \rmM_p^{(t)}}{\sqrt{d_k}} \right)\rmV_i^{(t)},
\end{equation*}
where $t = (\mathrm{type},l,m), \rmQ_i^{(\mathrm{type},l,m)} \in \R^{N\times d_k}$, $\rmK_i^{(\mathrm{type},l,m)} \in \R^{M\times d_k}$, $\rmV_i^{(\mathrm{type},l,m)} \in \R^{N\times d}$, $d_k$ and $d$ are hidden dimensions. Note that the only difference between $\widetilde{\rmA}_{i,p}^{(\mathrm{type},l,m)}$ and $\rmA_i^{(\mathrm{type},l,m)}$ is that in the calculation for the latter we replace $\rmM_p^{(\mathrm{type},l,m)}$ with the zero matrix. Also, note that our pruning mechanism only reduces the computational graph, and \textit{does not affect the number of parameters of the model}. 

An interpretation of inducing the mask $\rmM_p$ onto our computational graph is that we constrain the inductive bias of the attention mechanism by adding the inductive bias of \textit{whether a pair of source--target positions correlate}, which we learn from the training dataset by inspecting the outputs of a pre-trained model. Observing whether a pair of positions correlate in the training dataset would allow us to infer a positional inductive bias from the dataset itself that we can incorporate during training. Therefore, we propose the following algorithm.

\paragraph{Attention Pruning (AP).}
Choose a model $F$ with attention matrices in its computational graph and a percentage $p$ for pruning. Then follow the steps below:
\begin{enumerate}
\item Train the model $F$, initialized with weights $\vtheta$, on the training set $\train=\{(x_i,y_i)\}_{i=1,\dots,n}$ via validation on the $\valid$ split to obtain optimized weights $\vtheta^*$ and then compute the accuracy (or any other desired evaluation measure)
\item For each possible tuple $(\mathrm{type}, l, m)$, calculate the average $\overline{\rmA}^{(\mathrm{type}, l, m)}$ from the attention matrices \[\rmA_i^{(\mathrm{type},l,m)} \equiv \rmA_i^{(\mathrm{type},l,m)}(x_i;\vtheta^*)\]
for $i=1,\dots, N$, i.e.,~generated by each training example in $\train$, using~\eqref{eq:accum}.
\item For each possible tuple $(\mathrm{type}, l, m)$, calculate the mask $\rmM_p^{(\mathrm{type},l,m)}$ from $\overline{\rmA}^{(\mathrm{type}, l, m)}$ using~\eqref{eq:mask}, then obtain a new model $F'$ from $F$ by replacing the attention matrices with $\widetilde{\rmA}_{i,p}^{(\mathrm{type},l,m)}$ as in Section~\ref{sec:ap}.
\item Train $F'$ on $\train$ 
and compute the accuracy $a^{\mathrm{pruned}}_{\mathrm{test},p}$ on $\test.$
\end{enumerate}

Our method is inspired by the Lottery Ticket Hypothesis~\citep{lottery}. However, except from sharing Step 1, our AP method diverges significantly from other methods inspired by the Lottery Ticket Hypothesis, such as \citep{Yu2020Playing}, as we study sparseness not in the parameters of the neural network, but in the attention patterns of the model on a fixed dataset. Such a distinction is important, because while the sparseness of the weight matrices is usually not interpretable, the sparseness of the attention patterns is, as it has been observed in the literature~\citep{kovaleva-etal-2019-revealing,aless2020fixed,longformer}.

Given our method and the above motivation we now ask the following: (1) Can AP reveal a global sparseness for attention patterns? 
(2) Can that sparseness be deployed for efficient inference?
(3) Can AP yield an insight into the sparseness properties of different attention types? We answer positively all these questions in the following experimental sections.

\section{Language Modelling Experiments}
\label{sec:lm}

We first test pruning attention matrices on the WikiText-103~\citep{wt103} language modelling task. We use the Transformer-XL base architecture \citep{Dai2019TransformerXLAL}, which adds recurrence to Transformer models by caching self-attention hidden states between text segments. 

\begin{table}[t!]
  \small
  \centering
  \begin{tabular}{cc}
    \toprule
    \bf $p$ (\%) & \bf Perplexity  \\
    \cmidrule(r){1-2}
    0 & $24.157$  \\
    20 & $24.157$ \\
    40 & $24.214$  \\
    60 & $24.566$  \\ 
    80 & $25.115$ \\ 
    90 & $26.011$ \\
    \bottomrule
  \end{tabular}
  \caption{\rev{Attention Pruning for Transformer-XL-base trained on WikiText-103 can sparsify 90\% of its attention patterns, while maintaining good performance.}}
  \label{table:transformer_xl_results}
\end{table}

\begin{figure}[t!]
    \centering    \includegraphics[width=\linewidth]{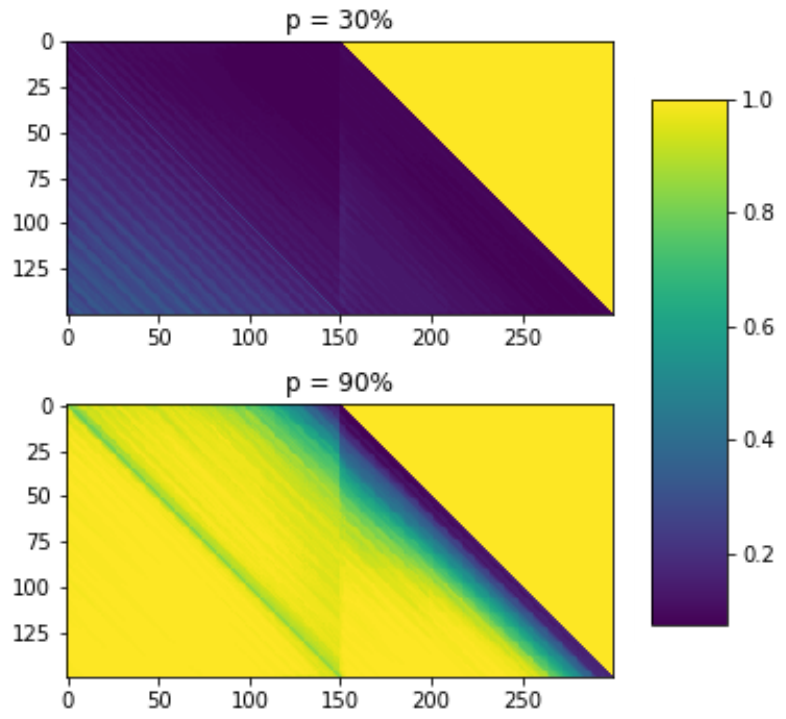}
    \caption{\rev{Transformer-XL pruning masks (binary valued) averaged over all layers and attention heads for $p \in \{30\%, 90\%\}$. AP prunes entries in the left half (attention to past sequences) more aggressively than the conventional self-attention entries in the right half. Note that the right half also has an auto-regressive mask.}}
    \label{fig:transformer-xl}
\end{figure}

We can see in Table~\ref{table:transformer_xl_results} and Figure~\ref{fig:visualization_ap}(a) that we can prune over 80--90\% of the attention entries and still maintain good performance on language modelling tasks. \rev{Figure~\ref{fig:transformer-xl} shows Transformer XL's prune masks, averaged over all its layers and heads, when using $p = 30\%$ or $p =  90\%$. We speculate that attention pruning performs so well here because it enables Transformer-XL to pay attention to long sequences only when the distant past is actually relevant.}

\section{Machine Translation Experiments}
\label{sec:mt_experimnets}
We further test our method on encoder-decoder Transformer models using various translation tasks and the implementation by \citet{ott2019fairseq}. In all experiments, we use half-precision floating point (FP16). We run two sets of experiments on the WMT17 English--German (en-de) and the IWSLT14 German--English (de-en) machine translation tasks following the default base Transformer 
architectures from \citet{ott2019fairseq} (\texttt{transformer\_iwslt\_de\_en} and \texttt{transformer\_wmt\_en\_de}, respectively), and we evaluate on the best model, measured by the highest validation-split BLEU score in the case of IWSLT14 de-en, and by the lowest validation loss in the case of WMT17 en-de.

\rev{
We prune all three types of attention ---\emph{self-attention-encoder}, \emph{self-attention-decoder} and  \emph{encoder-decoder}---, and we run five experiments per translation task, for $p\in \{ 20, 40, 50, 60, 80\}$, for both IWSLT14 de-en and WMT17 en-de.

\begin{table}[t!]
    \centering
    \small
    \begin{tabular}{cccccc}
        \toprule
         $p$ (\%) & \bf 0 & \bf 20  & \bf 40 
         & \bf 60 & \bf 80  \\
        \cmidrule(r){1-6}
          IWSLT14   & 34.94 & 32.08 &24.16 
          & 14.18 & 5.00 \\
        \cmidrule(r){1-6}
           WMT17  & 26.73 & 23.04 & 4.03 
           & 1.28 & 0.30\\
        \bottomrule
    \end{tabular}
    \caption{\rev{\newnewrev{BLEU scores} results from the IWSLT14 de-en and WMT17 en-de translation tasks. We can see that pruning all types of attention mechanism leads to a fast drop in performance.}}
    \label{table:fairseq_all}
\end{table}

Table~\ref{table:fairseq_all} summarizes the pruned models performance for the two translation tasks. We can see that AP performs significantly worse here than it did for the language modeling task above. \newrev{This suggest that a finer-grained analysis on pruning specific attention types, with which we proceed below.}
}

 \subsection{Attention Type Analysis} 
\label{subsec:type_analysis}

We now look at whether the three types of attention mechanism require specialized pruning. In order to isolate the effects of AP, we first prune only one of the three types of attention mechanisms. The results are summarized in Tables \ref{table:fairseq_iwslt_de-en_encOnly} and \ref{table:fairseq_wmt_en-de_encOnly}.

\begin{table}[t!]
  \small
  \centering
  \begin{tabular}{cccc}
    \toprule
    \bf $p$ (\%) & \bf Self-Enc  &  \bf Self-Dec & \bf Cross  \\

    \cmidrule(r){1-4}
    \cmidrule(r){1-4}
    
    0 & \multicolumn{3}{c}{34.94} \\
    
    \cmidrule(r){1-4}

    20 & 34.53 & 34.94 &  33.50 \\

    40 & 33.70 & 34.94 & 24.38  \\
    
    50 & 33.56 & 35.08&  22.60  \\
    
    60 & 33.68 & 34.91&  15.08  \\
    
    80 & 33.67 & 34.90&  6.39  \\

    \bottomrule
  \end{tabular}
  \caption{Results on the IWSLT14 de-en translation task when pruning each type of attention mechanism. Pruning cross-attention connections sharply hurts the model performance.}
   \label{table:fairseq_iwslt_de-en_encOnly}
\end{table}

\begin{table}[t!]
  \small
  \centering
  \begin{tabular}{cccc}
    \toprule
    \bf  $p$ (\%) & \bf Self-Enc  &  \bf Self-Dec & \bf Cross  \\

    \cmidrule(r){1-4}
    \cmidrule(r){1-4}
    
    0 & \multicolumn{3}{c}{26.73} \\
    
    \cmidrule(r){1-4}

    20 & 26.21 & 26.73 & 21.68  \\

    40 & 26.60 & 26.73 & 3.72   \\
    
    50 & 26.60 & 26.51 & 2.46 \\
    
    60 & 26.15 & 26.61 & 1.60   \\
    
    80 & 26.11 & 23.56  & 0.69  \\

    \bottomrule
    
  \end{tabular}
  \caption{Results obtained on the WMT17 en-de translation task when pruning each type of attention mechanism. Pruning up to 80\% of encoder self-attention connections results in a minimal drop of model performance.}
  \label{table:fairseq_wmt_en-de_encOnly}
\end{table}

\begin{figure*}[t!]
\centering
\begin{subfigure}{.31\linewidth}
\includegraphics[width=\linewidth]{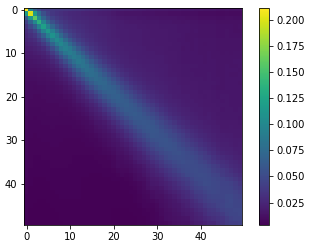}
\caption{}
\label{fig:average_iwslt14_enc_dec_entmaxFalse}
\end{subfigure}
\begin{subfigure}{.31\linewidth}
\includegraphics[width=\linewidth]{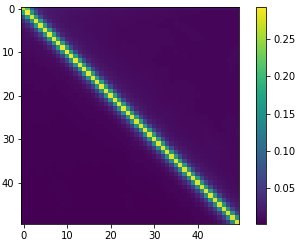}
\caption{}
\label{fig:average_iwslt14_self_enc_entmaxTrue}
\end{subfigure}
\begin{subfigure}{.31\linewidth}
\includegraphics[width=\linewidth]{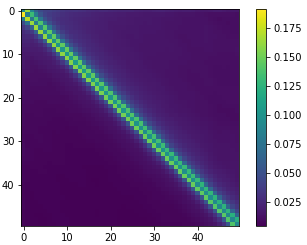}
\caption{}
\label{fig:average_iwslt14_self_enc_entmaxFalse}
\end{subfigure}
\caption{IWSLT14 de-en train dataset attention patterns: (\subref{fig:average_iwslt14_enc_dec_entmaxFalse}) cross-attention with variable context window, (\subref{fig:average_iwslt14_self_enc_entmaxTrue}) encoder self-attention with 1.5-entmax activation for sharper patterns, and (\subref{fig:average_iwslt14_self_enc_entmaxFalse}) encoder self-attention with constant context window.}
\label{fig:fused_iwslt14_patterns}
\end{figure*}

For both datasets, we clearly see the same trend: the Transformer models are more sensitive to pruning cross-attention patterns than to removing self-attention, which is in line with what was reported in previous work \rev{\cite{you-etal-2020-hard}. Figures~\ref{fig:average_iwslt14_enc_dec_entmaxFalse} and \ref{fig:average_iwslt14_self_enc_entmaxFalse}
 represent the average attention patterns observed on the IWSLT14 de-en dataset for cross-attention and the encoder's self-attention mechanism, respectively. A direct comparison between the two suggests that cross-attention mechanisms are more brittle, which is in part because they exhibit variable context windows. This makes sense since in this case the queries and the keys are generated from different sequences. 
 }

\subsection{Pruning Self-Attention Only}

We prune the two types of self-attention patterns (\emph{self-attention-encoder} and \emph{self-attention-decoder}) for $p \in \{ 20, 40, 50, 60, 80\}$. The results are shown in Table~\ref{table:only_self_fairseq}.

\begin{table}[t!]
  \centering
  \small
  \begin{tabular}{ccc}
    \toprule
    {\bf $p$ (\%)} & {\bf IWSLT14 de-en} & {\bf WMT17 en-de} \\
    \cmidrule(r){1-3}
    0 & 34.94   &  26.73 \\
    20 & 34.92  &  26.21 \\
    40 & 33.70  &  26.60 \\
    50 & 33.68  &  26.19 \\ 
    60 & 33.64  &  26.44 \\ 
    80 & 33.81  &  21.88 \\ 
    \bottomrule
  \end{tabular}
  \caption{\rev{BLEU scores for both translation tasks when pruning self-attention mechanisms. These are more robust under AP, even when the encoder and the decoder attentions are simultaneously pruned.}}
  \label{table:only_self_fairseq}
\end{table}

In agreement with \ref{subsec:type_analysis}, we find that we can prune large percentage of self-attention connections while maintaining good BLEU scores. The average attention patterns among all encoder self-attention heads for IWSLT14 are shown in Figure~\ref{fig:average_iwslt14_self_enc_entmaxFalse}. We can see that they are indeed sharper than in the case of cross-attention, and we show that a fixed window encodes the relevant contextual information for processing an input token. Moreover, we observe that numerous attention heads encode very similar patterns, in agreement with \cite{voita2019}.

The results for WMT17 en-de (see Figure~\ref{fig:visualization_ap}(b), Table ~\ref{table:only_self_fairseq}) are particularly encouraging. We can prune over 60\% of the self-attention entries, while losing less than 1 BLEU point absolute. \rev{This suggests that AP is particularly robust when used with large datasets, possibly because we use summary statistics.}

\section{Sparse Normalization}
\label{sec:sparsenorm}

\rev{One disadvantage of our pruning method is that we use a single pruning percentage $p$ for all attention patterns within a Transformer layer. Translation models are sequence-to-sequence models from a source fragment of length $M$ to a target fragment of length $N$, where $M$ and $N$ are \textit{not invariant} across a dataset. Because of this, we either leave extraneous noisy connection between tokens in shorter sequences or we lose important information when modeling longer sequences. }

This problem would be ameliorated if we masked attention mechanisms according to patterns sharper than those produced by the softmax activation function. Therefore, we turn to the $\alpha$-entmax activation, which was used in \citep{martins_adapt_sparse} and \citep{martins_sparse-seq2seq}, and which encourages sparseness in the attention patterns. We repeated all IWSLT14 de-en experiments from Section~\ref{sec:mt_experimnets} using $\alpha = 1.5$. Figure~\ref{fig:average_iwslt14_self_enc_entmaxTrue} shows the average observed attention pattern among the encoder self-attention heads in a model that uses 1.5-entmax as its normalizing activation function. We can see that they are indeed sharper than the analogous ones, \rev{which we presented} in Figure~\ref{fig:average_iwslt14_self_enc_entmaxFalse}.

\begin{table}[t!]
  \centering
  \small
  \begin{tabular}{cccccc}
    \toprule
    \bf $p$ (\%) &  \multicolumn{4}{c}{\bf BLEU score}  \\
    
    \midrule

     & Enc & Dec &  Cross  &  Enc+Dec  \\
    
    \midrule

    0 & \multicolumn{4}{c}{34.93}  \\
    20 & 34.38 & 34.93 & 32.02 & 34.55 \\
    40 & 34.76 & 34.93 & 24.92 & 34.60\\
    50 & 34.33 & 34.93 & 16.03 & 34.37\\
    60 & 33.77 & 35.05 & 10.97 & 33.91\\
    80 & 33.17 & 34.93 & 4.38  & 33.21\\
    \bottomrule
  \end{tabular}
  \caption{\rev{Experiments on IWSLT14  de-en using 1.5-entmax. We can see that pruning self-attention mechanisms maintains good performance at higher sparseness percentages than those induced by 1.5-entmax alone. This is in agreement with the trend observed in Section~\ref{subsec:type_analysis}, where we saw that pruning cross-attention patterns yields a sharp drop in performance.}}
  \label{table:all_entmax_fairseq_iwslt_de-en}
\end{table}

Table~\ref{table:all_entmax_fairseq_iwslt_de-en} demonstrates how attention pruning performs in conjunction with 1.5-entmax on the IWSLT14 de-en translation task. We would like to note that the behavior in \ref{subsec:type_analysis} regarding cross-attention's lack of robustness to pruning is even more pronounced now.

\section{AP with BERT on GLUE}
\label{sec:bert}

In order to analyze the sparseness of BERT on GLUE tasks~\citep{wang2018glue}, we study the vanilla BERT architecture \texttt{bert-base-cased}, and we apply our AP method on top of it. 
We train on the standard GLUE tasks with searched hyperparameters listed in Appendix~\ref{sec:glue}.

\begin{table*}[th]
  \small
  \centering
  \begin{tabular}{ccccccccccc}
    \toprule
    \multicolumn{11}{c}{\bf Accuracy (\%)} \\
    \midrule
    \multicolumn{1}{c}{\bf $p$ (\%)} &
    \multicolumn{1}{c}{\bf MNLI-m} & 
    \multicolumn{1}{c}{\bf MNLI-mm} & 
    \multicolumn{1}{c}{\bf QNLI} &
    \multicolumn{1}{c}{\bf QQP} &
    \multicolumn{1}{c}{\bf RTE} &
    \multicolumn{1}{c}{\bf SST-2} &
    \multicolumn{1}{c}{\bf MRPC} &
    \multicolumn{1}{c}{\bf CoLA} &
    \multicolumn{1}{c}{\bf STS-B} & 
    \multicolumn{1}{c}{\bf Average}\\
    \cmidrule(r){1-1}
    \cmidrule(r){2-2}
    \cmidrule(r){3-3}
    \cmidrule(r){4-4}
    \cmidrule(r){5-5}
    \cmidrule(r){6-6}
    \cmidrule(r){7-7}
    \cmidrule(r){8-8}
    \cmidrule(r){9-9}
    \cmidrule(r){10-10}
    \cmidrule(r){11-11}
    0 & 84.07 & 83.44 & 90.91 & 87.51 & 65.7 & 91.97 & 88.77 & 57.78 & 88.39 & 82.06 \\ 
    20 & 84.00 & 83.42 & 89.72 & 86.37 & 64.44 & 91.11 & 87.12 & 55.99 & 87.34 & 81.06 \\
    40 & 83.42 & 83.70 & 88.76 & 84.73 & 62.82 & 89.91 & 84.45 & 52.33 & 86.48 & 79.62 \\
    50 & 83.32 & 82.87 & 87.81 & 83.84 & 62.09 & 89.05 & 83.08 & 48.56 & 85.28 & 78.43 \\
    60 & 82.54 & 81.98 & 87.19 & 83.10 & 61.37 & 88.82 & 82.04 & 45.05 & 81.70 & 77.09 \\
    80 & 79.29 & 78.64 & 82.37 & 81.32 & 57.22 & 84.52 & 78.57 & 34.80 & 65.89 & 71.40 \\
    90 & 75.40 & 75.23 & 77.23 & 77.45 & 49.46 & 80.56 & 79.41 & 20.28 & 51.39 & 65.16 \\

    \bottomrule
  \end{tabular}
  \caption{BERT on GLUE. Attention Pruning reduces the attention computations by tens of percentage points, while maintaining comparable performance.
  We report Spearman correlation for STS-B, F1 for QQP and MRPC, and accuracy for the rest. The reported results are the median from five reruns on Dev. \label{table:bert}}
  
\end{table*}

In Table~\ref{table:bert} and Figure~\ref{fig:visualization_ap}(c), we present the results of our experiments. Generally, we observe that we can perform attention pruning on BERT heads, while maintaining the performance. For example, AP 
\rev{with $p=20$ loses only 1.00} points absolute on average
compared to the 
baseline $p=0$ model. Another important observation is that, for some GLUE tasks, we can prune even more than 50\% of the attention weights while maintaining competitive scores, e.g., for MNLI, \rev{QNLI, QQP, SST-2,} and SST-B.
For the remaining tasks ---RTE, MRPC and CoLA---, we maintain reasonable performance by pruning a sizable fraction of the attention weights.

\rev{
\section{Why Use Tailored Attention Masks}
\label{sec:ood_random}

We explore the importance of using attention masks tailored to specific datasets by comparing against two other scenarios: (\emph{i})~prune random entries in attention patterns, and (\emph{ii})~prune using an attention mask learned on a different dataset. 

\subsection{Machine Translation}

We  prune  self-attention  mechanisms in encoder--decoder  models  trained  with  softmax  on  the  IWSLT14 translation tasks for $p \in \{20,40,50,60,80\}$ using either random or out-of-distribution attention masks, and we compare to the baseline results above. For the out-of-distribution experiments, we generate attention masks on IWSLT14 en-de. 
Tables \ref{table:random} and \ref{table:ood_translation} show the results. We note that data-informed masks outperform random pruning by a large margin, especially as the percentage $p$ increases. However, using attention masks gathered for the other translation direction does not meaningfully influence our results. We speculate that this is because we only look at self-attention patterns, which in the case of translation are very sharp and exhibit a constant context window.

\begin{table}[t]
    \centering
    \small
    \setlength{\tabcolsep}{4pt}
    \begin{tabular}{ccccccc}
        \toprule
         \bf $p$ (\%) & \bf 0 & \bf 20  & \bf 40 & \bf 50 & \bf 60 & \bf 80  \\
        \cmidrule(r){1-7}
          AP & 34.94 & 34.92 & 33.70 & 33.68 &  33.64&  33.81 \\
       \cmidrule(r){1-7}
          Rand. & 34.94 & 32.51& 30.66& 27.96 & 26.56 &5.93 \\
        \bottomrule
    \end{tabular}
    \caption{Results for Machine Translation with IWSLT14 de-en task. \rev{Pruning random attention entries yields sharp drop in performance as the pruning percentage $p$ increases. AP yields up to $80\%$ sparseness, while maintaining good performance.}}
    \label{table:random}
\end{table}

\begin{table}[t]
    \centering
    \small
    \setlength{\tabcolsep}{4pt}
    \begin{tabular}{ccccccc}
        \toprule
         \bf $p$ (\%) & \bf 0 & \bf 20  & \bf 40 & \bf 50 & \bf 60 & \bf 80  \\
        \cmidrule(r){1-7}
          base & 30.57 & 30.37 & 27.61 & 27.28 &  27.51 &  27.46 \\
          ood & 30.57 & 30.44& 27.46 & 27.60 & 27.56 & 27.46 \\
        \bottomrule
    \end{tabular}
    \caption{\newrev{Results from the IWSLT14 en-de translation tasks. For simple attention patterns, such as those in self-attention mechanism for translation tasks, AP is robust under distributional shifts.}}
    
    \label{table:ood_translation}
\end{table}

\subsection{BERT}
We perform out-of-domain experiments by training GLUE tasks with AP using attention patterns from all GLUE tasks. In Figure~\ref{fig:bert}, we show the relative difference in accuracy for our experiments on four GLUE datasets. Note that the columns for SST-2 and CoLA show significantly negative relative accuracy, which means that the attention patterns from these datasets are not useful for the majority of the GLUE tasks. This is in line with the nature of the datasets: SST-2 and CoLA are two tasks in GLUE that do not involve pairs of sentences. Particularly detrimental to the performance is using CoLA masks on the STS-B dataset (black entry in Figure~\ref{fig:bert}). Our experiment in the figure is for $p=40$, but we observed a similar pattern for a variety of pruning percentages.

\begin{figure}[t!]
    \centering    \includegraphics[width=\linewidth]{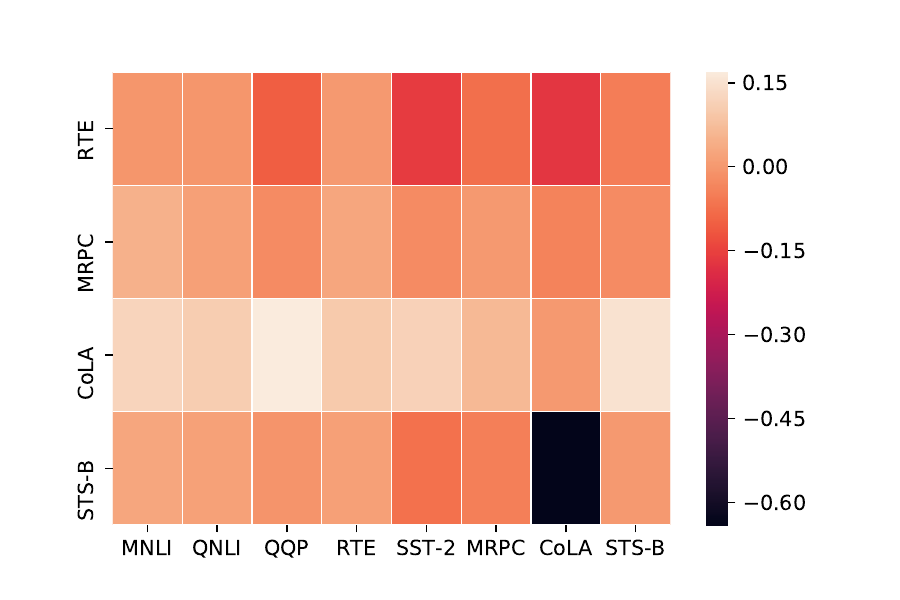}
    \caption{\rev{Relative accuracy when a GLUE task (indicated in the rows: STS-B, CoLA, MRPC, RTE) is trained with AP ($p=40$) using the attention patterns of GLUE tasks (indicated in the columns: all GLUE tasks). The relative accuracy is computed so that the in-domain experiment is zero, and the out-of-domain experiments show deviations in accuracy.}}
    \label{fig:bert}
\end{figure}
}

\section{Towards Efficient Hardware Implementation of Attention Pruning}
\label{sec:hardware}
\subsection{\newrev{MACs}}
\label{sec:MACS}

In order to quantify the computational advantage we obtain from pruning, we estimate the number of Multiply-Accumulate Operations, or MACs for short~\citep{randel1971}, executed by a simple attention mechanism during the forward pass of the model.

Let the input be a tensor $x\in \mathbb{R}^{B \times N \times d}$, where $B$ is the batch size, $N$ is the number of tokens in a sequence, and $d$ is the number of embedding dimensions. In Appendix~\ref{sec:appendix} we show that with AP we achieve a reduction of
\begin{equation}
\label{eq:reduction}
\text{fraction of MACs} = \frac{4d + (2-p)N}{4d + 2N}.
\end{equation}
\newrev{This is particularly helpful for large $N \gg d,$ suggesting that AP might be computationally beneficial in applications with long input sequences such as summarization or question answering.}

\subsection{\newnewrev{Bert Benchmark}}

\newrev{
We demonstrate empirical gains using block-sparse GPU kernels. \citet{gray2017} first implemented efficient sparse matrix multiplication operations and applied them to both dense and convolutional layers. \citet{Child2019GeneratingLS} extended this work to attention mechanisms with certain pre-determined sparseness patterns. \citet{10.1145/3315508.3329973} introduced Triton, a language and compiler used to generate optimized GPU code that allows for higher design flexibility than Pytorch. We use the DeepSpeed\footnote{https://www.deepspeed.ai} implementation of sparse attention, which requires efficient sampled dense-dense and sparse-dense multiplications as well as softmax operations.

Since in order to obtain performance gains we need to use sparseness patterns structured around blocks, as a proof-of-concept, we turn our attention to question answering applications and apply AP to the Stanford Question Answering Dataset (SQuAD) \citep{2016arXiv160605250R} and use a blocksize of 16. We choose SQuAD for hardware benchmarking as the sequences are longer (up to 384 tokens) than those in the GLUE benchmark. We fine-tune BERT on  a  single  GeForce GTX 1080 GPU for two  epochs  with  a  learning  rate  of  3e-5  and a batch size of 4. The length of the sequences is capped at 384 tokens. At test time, we use a batch size of 256.

\begin{table}[t!]
  \centering
  \small
  \begin{tabular}{clll}
    \toprule
     \textbf{Kernel} &
     \textbf{Pruning} & \multicolumn{2}{c}{\textbf{Efficiency}} \\
     \cmidrule(r){2-2}
     \cmidrule(r){3-4}
     & $p$\textbf{ (\%)}  &  \textbf{Time (s)}  & \textbf{Memory (GB)} \\
     \cmidrule(r){1-1}
     \cmidrule(r){2-2}
     \cmidrule(r){3-3}
     \cmidrule(r){4-4}
    Pytorch & \grayc{0} & \grayc{95.80} & \grayc{6.24}  \\
    \midrule
    Triton & 0 & 95.41 & 6.85 \\
    Triton & 90 &  86.44 \improve{9.4\%} & 5.00 \improve{27\%} \\
    \bottomrule
  \end{tabular}
  \caption{Proof-of-concept empirical gains from using AP inference on SQuAD. Memory tracks the GPU.}
  \label{table:squad}
\end{table}

The results in Table~\ref{table:squad} suggest that we can empirically reduce both inference time and GPU memory consumption using AP and sparse kernels. At the same time, we do not compromise performance noticeably: Pytorch and Triton kernels for $p=0$ yield 81.02 Exact and 88.63 F1 scores, while AP with Triton for $p=90$ yields 79.62 Exact and 87.32 F1 scores.
We would like to underscore the 27\% reduction in memory, in particular, as memory limitations usually prevent utilization of attention-based models when hardware is constrained. Thus AP is especially promising in this context.

\newnewrev{

\subsection{Llama2 7B Benchmark } 

We evaluate our pruning method on the Llama2 7B model using the Wikitext-2 dataset, with a focus on language modeling \footnote{https://github.com/irugina1/llama-attention-pruning}. We use a context length and stride of 4096. Attention pattern statistics are gathered from the training split, and evaluation is performed on the test split. Due to computational constraints, we do not fine-tune the model weights. Given this limitation, we concentrate on moderate pruning percentages. This approach demonstrates how memory improvements can be scaled to recent large language models even with low batch sizes; all experiments are conducted with a batch size of one.

\begin{table}[h!]
\centering
\begin{tabular}{l l l}
\toprule
 & \textbf{Pytorch} & \textbf{Triton} \\ \hline
\textbf{Pruning \%} & 0 & 60 \\
\textbf{Memory (GB)} & 19.51 & 16.88 \\
\textbf{Perplexity} & 6.48 & 6.72 \\ \bottomrule
\end{tabular}
\caption{Memory efficiency in Llama2 7B model: Pruning cuts forward pass memory from 19.51 GB to 16.88 GB, over a 13.92 GB baseline to load the model, with minimal perplexity increase.}
\label{table:llama2}
\end{table}

We summarize our results in Table \ref{table:llama2}. Loading the model onto the GPU with half-precision weights requires 13.92 GB of memory. We managed to reduce the additional memory requirements for a forward pass from 5.58 GB to 3.07 GB, achieving a reduction of over 40\%, with only a minimal increase in perplexity.

}

}

\section{Conclusion and Future Work}
\label{sec:conclusion}
We introduced Attention Pruning, a novel method that leverages data-informed sparseness for pruning attention. Through controlled experiments on a variety of tasks (from language modeling to machine translation and GLUE tasks predictions), we showed that our method prunes most computations using pre-computed attention patterns while maintaining, or even improving, performance. Our application to seq2seq tasks enabled us to study attention patterns in self- and cross-attention, revealing key distinctions between the two.

In future work, we aim to assess our method on other models, NLP tasks, and datasets of various sizes, as well as optimize Attention Pruning for existing hardware. We believe co-design approaches for efficient sparse kernels and their successful utilization can enhance Attention Pruning's scalability. To encourage further co-design efforts, we are releasing our code to the research community.

\section{Limitations}
Our study required extensive experiments across three types of tasks, language modeling, machine translation and BERT fine-tuning. If one wants to reproduce our analysis from scratch they would need to rely on extensive GPU compute. However, we will release our code, in hopes to alleviate the work of researchers who would like to perform a similar study.

\section{Ethics Statement}
In this research, we introduce a novel method for pruning attention mechanisms in natural language processing models. While our primary goal is to reduce computational complexity and enhance model efficiency, we recognize the potential ethical implications of our work.

First, it is important to note that our developed pruning technique can lead to the creation of more efficient and less resource-intensive models. This has positive implications for sustainability, as it could help reduce the energy consumption and environmental footprint of large-scale NLP models.

However, we also acknowledge that more efficient models might contribute to the development and deployment of increasingly powerful NLP systems, which could be misused for malicious purposes, such as disinformation campaigns or automated harassment. Therefore, we encourage the research community to adopt responsible practices in the development and deployment of NLP models with attention pruning and to continuously evaluate the potential risks and societal impact of these technologies.

Moreover, as our pruning method is applied to pre-trained models, the potential biases embedded in the training data could still persist. Therefore, we emphasize the importance of addressing and mitigating biases in NLP models and training data to ensure that the resulting systems do not perpetuate harmful stereotypes or unfair treatment of certain social groups.

Finally, we commit to transparently sharing our code, as mentioned earlier, and findings to encourage further collaboration within the research community, fostering an open and responsible approach to the development and improvement of attention pruning techniques.

\bibliography{anthology,custom}
\bibliographystyle{lrec-coling2024-natbib}

\appendix

\section{Estimation of the MACs in Equation \ref{eq:reduction}}
\label{sec:appendix}
In order to process the input $x$, we need to follow this sequence of operations: (\emph{i})~Compute the \textit{key}, \textit{query}, and \textit{value} matrices; each of these is obtained by applying a fully connected transformation $\mathbb{R}^d \rightarrow \mathbb{R}^d$ and in total requires $3BNd^2$ MAC operations; (\emph{ii})~Compute attention scores $P$; we need to compute $Bh$ matrix multiplications of the form $X_1 \times X_2$ between matrices $X_1\in \mathbb{R}^{N \times d_k}$ and  $X_2 \in \mathbb{R}^{d_k \times N}$; this takes $BhN^2d_k$ MACs.  (\emph{iii})~Compute attention patterns: we need to calculate $Bh$ matrix products of the form $X_1 \times X_2$ between matrices $X_1\in \mathbb{R}^{N \times N}$ and  $X_2 \in \mathbb{R}^{N \times td_k}$; this costs us $BhN^2d_k$ MACs; (\emph{iv})~Apply a fully connected layer to compute the attention mechanism output, which takes $BNd^2$ MACs. Now let us consider what happens when a fraction $p$ ($0 \leq p \leq 1$) of the entries in the matrix $P \in \mathbb{R}^{B \times h \times N \times N} $ are pruned: in step 3 we only need to do a fraction $1-p$ of the work we used to and therefore, we achieve reduction of
\[
\text{fraction of MACs} = \frac{4d + (2-p)N}{4d + 2N}.
\]

\section{Details about GLUE}
\label{sec:glue}

We train on the following GLUE tasks: MNLI~\citep{williams2018broadcoverage}, QQP~\citep{qqp}, STS-B~\citep{cer-etal-2017-semeval}, SST-2~\citep{socher-etal-2013-recursive}, RTE~\citep{Bentivogli09thefifth}, MRPC~\citep{dolan-brockett-2005-automatically}, QNLI~\citep{rajpurkar2016squad}, CoLA~\citep{Warstadt_2019}. For each task, we fine-tune the BERT model on a single GPU for three epochs with a learning rate of 2e-5 and a batch size of 32. The maximum length of the input sequences is 128. Since the datasets are small, we run each of our experiments with five different random seeds in order to be able to capture the mean and the standard deviation of the results.

\end{document}